\documentclass{article}

\PassOptionsToPackage{numbers, compress}{natbib}

\usepackage[final]{neurips_2020}
\usepackage[utf8]{inputenc} 
\usepackage[T1]{fontenc}    
\usepackage[breaklinks=true,colorlinks,urlcolor=purple]{hyperref}
\usepackage{url}            
\usepackage{booktabs}       
\usepackage{amsfonts}     
\usepackage{nicefrac}       
\usepackage{microtype}     
\usepackage{graphicx}
\usepackage{enumitem}
\usepackage[table]{xcolor}

\usepackage{times}
\usepackage{epsfig}
\usepackage{amsmath}
\usepackage{amssymb}
\usepackage{caption}
\usepackage{subcaption}
\usepackage{multirow}
\usepackage{makecell}
\usepackage{array}
\usepackage{longtable}
\usepackage{colortbl}
\usepackage{xcolor}
\usepackage{adjustbox}
\usepackage{changepage}

\title{NAS-NeRF: Generative Neural Architecture Search for Neural Radiance Fields}

\author{Saeejith Nair$^{1}$ \quad Yuhao Chen$^{1}$ \quad Mohammad Javad Shafiee$^{1,2,3}$ \quad Alexander Wong$^{1,2,3}$ \\
			$^{1}$ Vision and Image Processing Research Group, University of Waterloo\\
			$^{2}$ Waterloo Artificial Intelligence Institute, Waterloo, ON\\
			$^{3}$ DarwinAI Corp., Waterloo, ON\\ 
{\tt\small {\{smnair, yuhao.chen1, mjshafiee, a28wong\}}@uwaterloo.ca}
}

\begin{document}

\maketitle

\begin{abstract}
Neural radiance fields (NeRFs) enable high-quality novel view synthesis, but their high computational complexity limits deployability. While existing neural-based solutions strive for efficiency, they use one-size-fits-all architectures regardless of scene complexity. The same architecture may be unnecessarily large for simple scenes but insufficient for complex ones. Thus, there is a need to dynamically optimize the neural network component of NeRFs to achieve a balance between computational complexity and specific targets for synthesis quality. We introduce NAS-NeRF, a generative neural architecture search strategy that generates compact, scene-specialized NeRF architectures by balancing architecture complexity and target synthesis quality metrics. Our method incorporates constraints on target metrics and budgets to guide the search towards architectures tailored for each scene. Experiments on the Blender synthetic dataset show the proposed NAS-NeRF can generate architectures up to 5.74$\times$ smaller, with 4.19$\times$ fewer FLOPs, and 1.93$\times$ faster on a GPU than baseline NeRFs, without suffering a drop in SSIM. Furthermore, we illustrate that NAS-NeRF can also achieve architectures up to 23$\times$ smaller, with 22$\times$ fewer FLOPs, and 4.7$\times$ faster than baseline NeRFs with only a 5.3\% average SSIM drop. Our source code is also made publicly available\footnote{Project website:  \url{https://saeejithnair.github.io/NAS-NeRF}}.

\end{abstract}

\section{Introduction}
Neural radiance fields (NeRFs)~\cite{mildenhall2020nerf} can enable photorealistic novel view synthesis of complex 3D scenes by using multilayer perceptrons (MLPs) to represent continuous volumetric radiance and density fields. However, converging to a sufficiently high-resolution representation is computationally demanding, as NeRF requires hundreds of millions of costly neural network queries per rendered image.  However, converging to a sufficiently high-resolution representation is computationally demanding, as NeRF requires hundreds of millions of costly neural network queries per rendered image. For example, Mildenhall et al. report that training NeRF~\cite{mildenhall2020nerf} on their Blender synthetic dataset takes 640k rays per image, with each ray requiring 256 sampled points that need to be passed through the fields in order to accumulate colour and density. This results in over 150 million network queries per rendered image, which severely limits deployability, especially on resource constrained platforms.

Recent advancements have succeeded in enhancing the efficiency of NeRF by employing techniques like caching~\cite{garbin_fastnerf_2021, muller_instant_2022}, baking~\cite{hedman_baking_2021}, tensor decomposition~\cite{chen_tensorf_2022},  efficient sampling~\cite{neff_donerf_2021}, spatial decomposition~\cite{rebain_derf_2021, reiser_kilonerf_2021, garbin_fastnerf_2021} or multi-resolution hash encodings~\cite{muller_instant_2022}. However, these methods typically adopt one-size-fits-all architectures, irrespective of scene complexity and target metrics, making them impractical for realistic deployment constraints. 

We propose NAS-NeRF, a generative architecture search strategy~\cite{wong_ferminets_2018} that tailors scene-specific architectures. Using the NAS-NeRF field cell as our core, we modularize the neural radiance field representation in terms of a cell~\cite{zoph_neural_2017}. This enables us to reduce neural network complexity to meet deployment constraints without sacrificing synthesis quality. Experiments demonstrate that NAS-NeRF generates architectures up to 23$\times$ smaller, 22$\times$ fewer FLOPs, and 4.7$\times$ faster than baseline NeRFs with only a 5.3\% average SSIM drop across scenes on the Blender synthetic dataset~\cite{mildenhall2020nerf}.

\begin{figure}[t]
    \includegraphics[width=\textwidth]{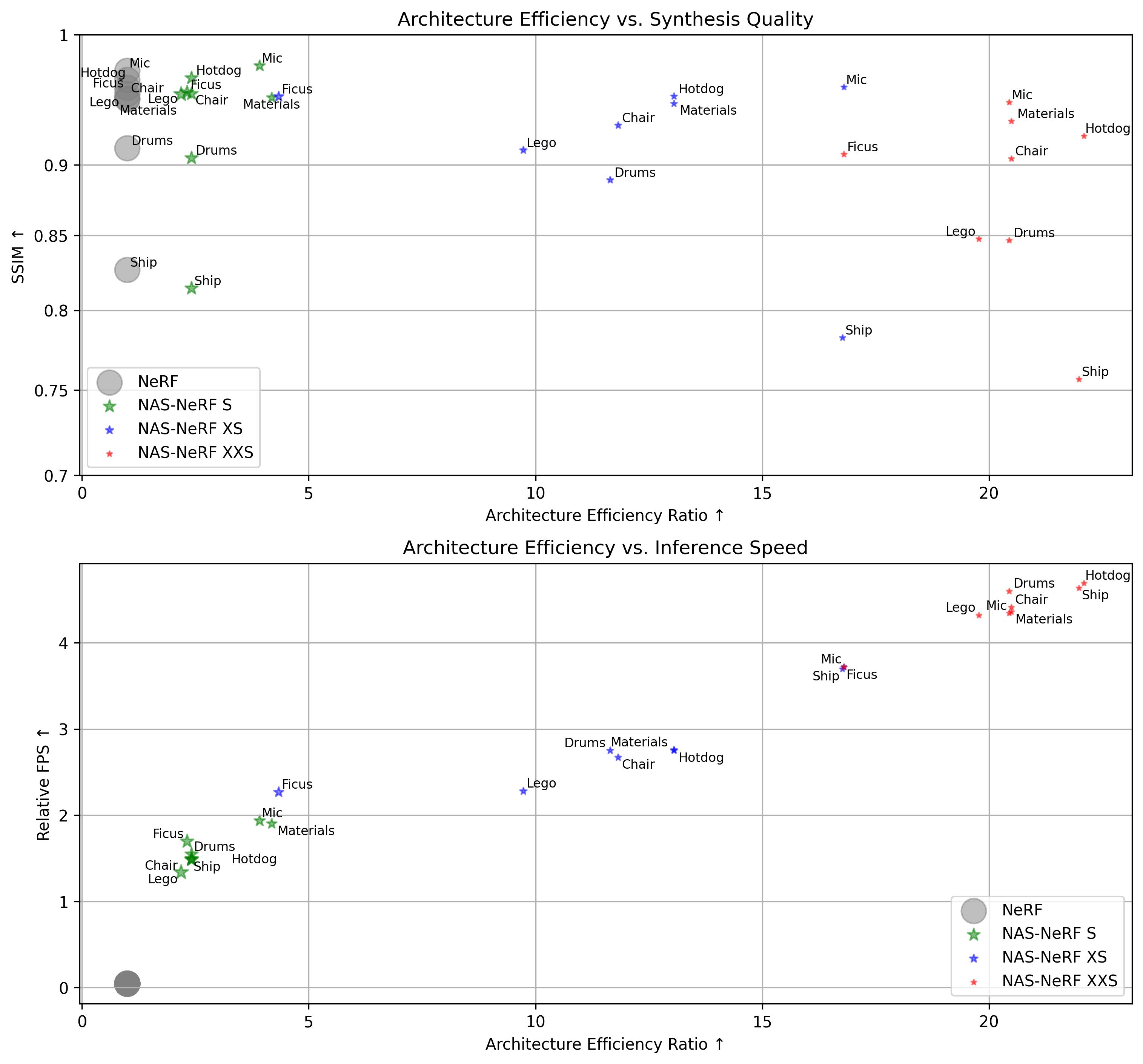}
\caption{Architecture efficiency ratio (a measure of generated architecture FLOPs relative to baseline NeRF) vs. (top) synthesis quality (SSIM)  and (bottom) inference speed; size $\propto$ parameter count.}
    \label{fig:efficiency}
\end{figure}

The remainder of this paper is organized as follows. Section 2 describes the methodology behind the creation of the proposed NAS-NeRF via generative neural architecture search, as well as a description of the resulting NAS-NeRF architectures. Section 3 describes the dataset used in this study, the training and evaluation setup, as well as the experimental results and complexity comparisons.

\section{Methods}

\subsection{NAS-NeRF Field Cell}
\begin{figure}[h]
    \centering
    \includegraphics[width=0.9\textwidth]{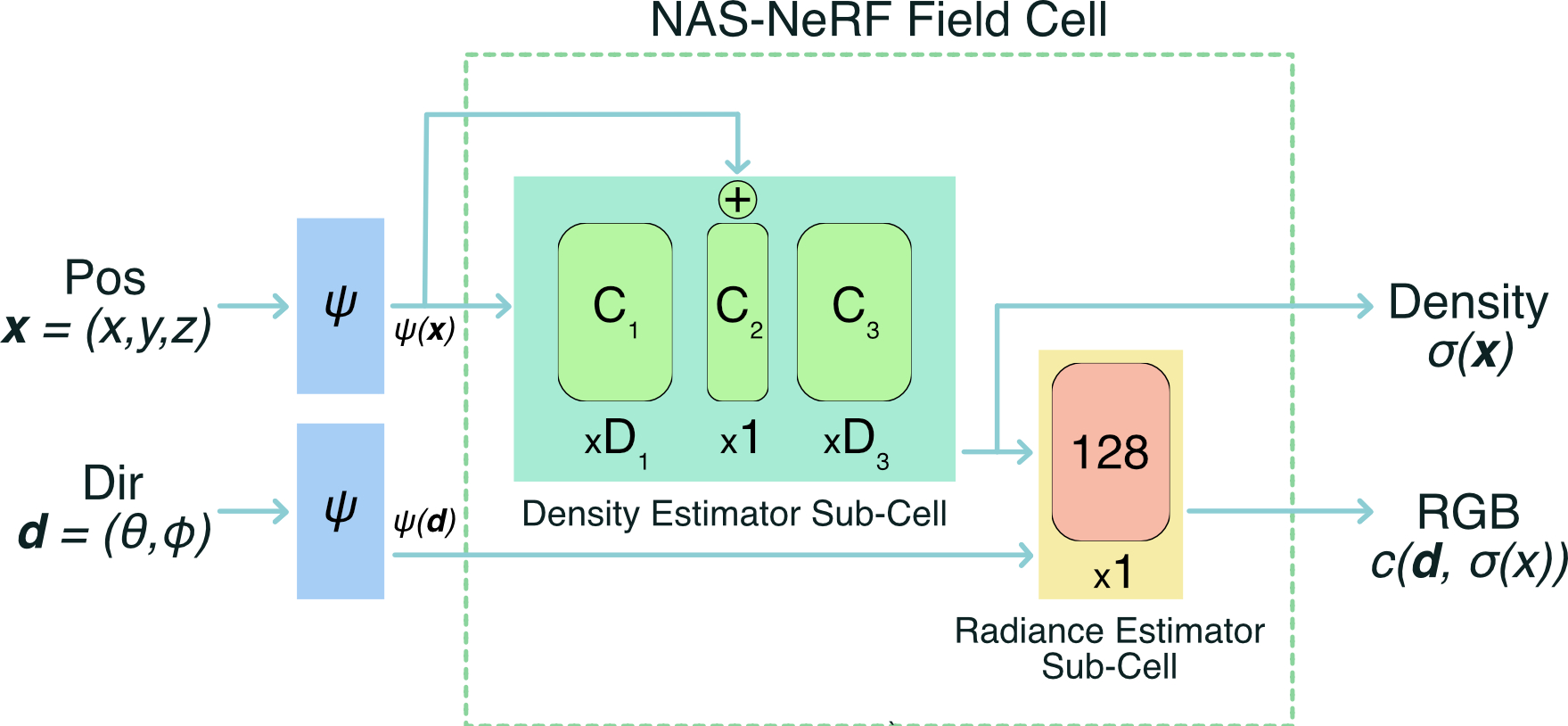}
\caption{High level overview of the NAS-NeRF field cell architecture. The NAS-NeRF pipeline comprises the original NeRF components~\cite{mildenhall2020nerf}, utilizing two NAS-NeRF field cells for coarse and fine hierarchical sampling. Each cell is parameterized by depth stages ($D_1$, $D_3$) and channels ($C_1$, $C_2$, $C_3$), where $\psi$ denotes the positional encoder for input coordinates or viewing directions. The nature of our parameterization ensures that the NAS-NeRF field cell can be plugged into most other NeRF methods, as our optimizations revolve entirely around the core network architectures. }
    \label{fig:cell}
\end{figure}

The NAS-NeRF field cell modularizes the classic NeRF field~\cite{mildenhall2020nerf} into stages to enable efficient architecture search. The classic NeRF field maps 3D spatial coordinates $\mathbf{x}$ and 2D viewing directions $\mathbf{d}$ to volume density $\sigma(\mathbf{x})$ and view-dependent emitted radiance $c(\mathbf{x}, \mathbf{d})$. This mapping is learned by training an 8-layer density estimator MLP and smaller single layer radiance estimator MLP. We focus our architecture search on the density estimator since it contributes the most parameters. As shown in Figure~\ref{fig:cell}, we decompose the density estimator into 3 stages with configurable parameters:

\begin{enumerate}
\item $D_1$ fully connected layers with $C_1$ channels
\item $D_2 = 1$ fully connected layer with $C_2$ channels and a skip connection concatenating to the input
\item $D_3$ fully connected layers with $C_3$ channels
\end{enumerate}

Decomposing the networks for neural radiance fields into standalone blocks enables reformulating NeRF architecture search as learning specialized cells, with the NeRF field acting analogously to an optimizable cell~\cite{zoph_neural_2017}. Furthermore, this allows viewing the overall NAS-NeRF architecture as two configurable field cells in series - one coarse field cell to learn global structure, and one fine field cell to learn high-frequency details. This modular parameterization focuses optimization on the core NeRF network while decoupling peripheral components like samplers or encoders. Our search strategy can thus function as a plugin applied to various NeRF methods by optimizing the architecture of the density estimator sub-cell.

\subsection{Generative Neural Architecture Search}
To discover NAS-NeRF architectures that balance efficiency and novel view synthesis quality, we employ a generative neural architecture search approach based on generative synthesis~\cite{wong_ferminets_2018}. Given operational constraints like compute budgets and target rendering metrics, generative synthesis executes an architectural exploration process to discover highly tailored architectures fitting the constraints. This discovery process can be formulated as a constrained optimization problem:
\begin{equation}
\mathcal{G}=\max_{\mathcal{G}}\mathcal{U}(\mathcal{G}(s)) \;\;\;\textrm{ subject to} \;\;\; 1_r(G(s))=1, \;\;\forall\in \mathcal{S}.
\label{eq:generative_synthesis}
\end{equation}
\noindent where the underlying objective is to learn an expression $\mathcal{G}(\cdot)$ that, given seeds $\{s|s \in S\}$, can generate network architectures $\{N_s|s \in S\}$ that maximizes a universal performance metric $U$ (e.g.,~\cite{wong_netscore_2019}) while adhering to operational constraints set by the indicator function $1_r(\cdot)$. This constrained optimization is solved iteratively through a collaboration between a generator $G$ and an inquisitor $I$ which inspects the generated network architectures and guides the generator to improve its generation performance towards operational requirements (see~\cite{wong_ferminets_2018} for details).

Although the NAS-NeRF search space is flexible, we enforce design constraints through $1_r$(·) in Eq.~\ref{eq:generative_synthesis} to achieve the desired balance between i) accuracy, ii) architectural complexity, and iii) computational complexity to yield low-footprint, compact NeRF architectures that are tailored for a target performance metric. Specifically, our constraints encourage:
\begin{enumerate}
\item Fields with both uniform and variable number of channels across different MLP stages.
\item Fields that uniformly expand such that stages of the MLP get progressively wider.
\item Meeting minimum scene-specific targets $T$ for performance metrics.
\end{enumerate}
 
\begin{table}[t]
\begin{adjustwidth}{-1.5cm}{-1cm}
\centering
\small
\begin{tabular}{c|c|cc|cc|c|c|c}
\toprule
Scene & Architecture & \multicolumn{2}{c|}{Coarse Field} & \multicolumn{2}{c|}{Fine Field} & FPS & Parameters (M) & FLOPs (G) \\
& & Depths  & Channels & Depths & Channels & (Speedup) & \\
\midrule
& NeRF & 8 $\times$  & 256 & 8$\times$ & 256 & 1$\times$ & 1.09 (1$\times$) & 574.14 (1$\times$) \\
\midrule
\midrule
Chair & NAS-NeRF S & [2, 1, 1] & [9, 11, 12] & [3, 1, 2] & [200, 207, 214] & 1.49$\times$  & 0.32 (3.46$\times$ ) & 237.57 (2.42$\times$ )\\
Chair & NAS-NeRF XS & [2, 1, 1] & [12, 12, 12] & [3, 1, 2] & [53, 57, 61] & 2.67$\times$  & 0.08 (14.33$\times$ ) & 48.56 (11.82$\times$ )\\
Chair & NAS-NeRF XXS & [2, 1, 1] & [9, 11, 12] & [2, 1, 1] & [16, 18, 20] & 4.41$\times$  & 0.05 (21.92$\times$ ) & 28.01 (20.49$\times$ )\\
\midrule
Drums & NAS-NeRF S & [2, 1, 1] & [9, 11, 12] & [3, 1, 2] & [200, 207, 214] & 1.55$\times$  & 0.32 (3.46$\times$ ) & 237.57 (2.42$\times$ )\\
Drums & NAS-NeRF XS & [2, 1, 1] & [20, 20, 20] & [3, 1, 2] & [53, 57, 61] & 2.75$\times$  & 0.08 (13.81$\times$ ) & 49.29 (11.65$\times$ )\\
Drums & NAS-NeRF XXS & [2, 1, 1] & [12, 12, 12] & [2, 1, 1] & [16, 18, 20] & 4.60$\times$  & 0.05 (21.82$\times$ ) & 28.08 (20.45$\times$ )\\
\midrule
Ficus & NAS-NeRF S & [2, 1, 1] & [12, 12, 12] & [3, 1, 2] & [214, 214, 214] & 1.70$\times$  & 0.33 (3.32$\times$ ) & 247.57 (2.32$\times$ )\\
Ficus & NAS-NeRF XS & [2, 1, 1] & [9, 11, 12] & [2, 1, 1] & [167, 174, 180] & 2.27$\times$  & 0.18 (5.99$\times$ ) & 132.33 (4.34$\times$ )\\
Ficus & NAS-NeRF XXS & [2, 1, 1] & [9, 11, 12] & [2, 1, 1] & [33, 36, 39] & 3.72$\times$  & 0.06 (18.94$\times$ ) & 34.17 (16.80$\times$ )\\
\midrule
Hotdog & NAS-NeRF S & [2, 1, 1] & [9, 11, 12] & [3, 1, 2] & [200, 207, 214] & 1.49$\times$  & 0.32 (3.46$\times$ ) & 237.57 (2.42$\times$ )\\
Hotdog & NAS-NeRF XS & [2, 1, 1] & [12, 12, 12] & [3, 1, 2] & [51, 51, 51] & 2.75$\times$  & 0.07 (15.51$\times$ ) & 43.98 (13.05$\times$ )\\
Hotdog & NAS-NeRF XXS & [2, 1, 1] & [12, 12, 12] & [2, 1, 1] & [9, 11, 12] & 4.69$\times$  & 0.05 (23.05$\times$ ) & 25.98 (22.10$\times$ )\\
\midrule
Lego & NAS-NeRF S & [2, 1, 1] & [100, 104, 109] & [3, 1, 2] & [214, 214, 214] & 1.34$\times$  & 0.39 (2.83$\times$ ) & 262.61 (2.19$\times$ )\\
Lego & NAS-NeRF XS & [2, 1, 1] & [12, 12, 12] & [4, 1, 3] & [64, 64, 64] & 2.28$\times$  & 0.09 (12.19$\times$ ) & 58.98 (9.73$\times$ )\\
Lego & NAS-NeRF XXS & [2, 1, 1] & [16, 18, 20] & [2, 1, 1] & [20, 20, 20] & 4.32$\times$  & 0.05 (20.64$\times$ ) & 29.03 (19.78$\times$ )\\
\midrule
Materials & NAS-NeRF S & [2, 1, 1] & [16, 18, 20] & [2, 1, 1] & [180, 180, 180] & 1.90$\times$  & 0.19 (5.74$\times$ ) & 137.17 (4.19$\times$ )\\
Materials & NAS-NeRF XS & [2, 1, 1] & [12, 12, 12] & [3, 1, 2] & [51, 51, 51] & 2.76$\times$  & 0.07 (15.51$\times$ ) & 43.98 (13.05$\times$ )\\
Materials & NAS-NeRF XXS & [2, 1, 1] & [9, 11, 12] & [2, 1, 1] & [16, 18, 20] & 4.36$\times$  & 0.05 (21.92$\times$ ) & 28.01 (20.49$\times$ )\\
\midrule
Mic & NAS-NeRF S & [4, 1, 3] & [56, 60, 64] & [2, 1, 1] & [180, 180, 180] & 1.93$\times$  & 0.23 (4.83$\times$ ) & 146.53 (3.92$\times$ )\\
Mic & NAS-NeRF XS & [2, 1, 1] & [9, 11, 12] & [2, 1, 1] & [33, 36, 39] & 3.72$\times$  & 0.06 (18.94$\times$ ) & 34.17 (16.80$\times$ )\\
Mic & NAS-NeRF XXS & [2, 1, 1] & [12, 12, 12] & [2, 1, 1] & [16, 18, 20] & 4.34$\times$  & 0.05 (21.82$\times$ ) & 28.08 (20.45$\times$ )\\
\midrule
Ship & NAS-NeRF S & [2, 1, 1] & [9, 11, 12] & [3, 1, 2] & [200, 207, 214] & 1.48$\times$  & 0.32 (3.46$\times$ ) & 237.57 (2.42$\times$ )\\
Ship & NAS-NeRF XS & [2, 1, 1] & [12, 12, 12] & [2, 1, 1] & [33, 36, 39] & 3.69$\times$  & 0.06 (18.86$\times$ ) & 34.23 (16.77$\times$ )\\
Ship & NAS-NeRF XXS & [2, 1, 1] & [9, 11, 12] & [2, 1, 1] & [12, 12, 12] & 4.63$\times$  & 0.05 (23.05$\times$ ) & 26.11 (21.99$\times$ )\\

\bottomrule
\end{tabular}
\caption{The architecture configuration and performance metrics of each generated NAS-NeRF is shown, along with the baseline NeRF~\cite{mildenhall2020nerf}. The architecture configurations for GEN-NeRF are listed as $[D_1, D_2, D_3]$ for the depth columns and $[C_1, C_2, C_3]$ for the channels columns. The architecture configuration for the baseline NeRF model is also listed as each field has an MLP with 8 layers and 256 channels. The FPS and Parameters columns show the frames per second and the number of parameters of each model, along with the speedup and architecture efficiency ratio relative to the baseline NeRF architecture.}
\label{tab:architecture_configs}
\end{adjustwidth}
\end{table}

Our experiments use SSIM~\cite{wang_image_2004} as the performance metric. For each scene, we sample 3 SSIM targets and generate 3 NAS-NeRF variants per scene (XXS, XS, S) that achieve different trade-offs between efficiency and quality. 

Determining suitable targets for a new scene is challenging, as we can't know \textit{a priori} what performance can be achieved. We circumvent this using an efficient heuristic which is fast and extensible to any scene or NeRF method. Specifically, we train two boundary architectures~$ A$, a maximum size architecture $A_{\textit{max}}$ and minimum size $A_{\textit{min}}$ ($> 23.2 \times$ fewer parameters than the baseline) for 16k iterations each (this takes 15-60mins on an Nvidia RTX A6000 GPU). Using their SSIM evaluations $\textit{SSIM}_{\textit{min}}$ and $\textit{SSIM}_{\textit{max}}$, we linearly interpolate targets $T_{\textit{XXS}}$, $T_{\textit{XS}}$, and $T_{\textit{S}}$ at 10\%, 50\%, and 90\% between the extremes. Although SSIM versus size may be non-linear along the true pareto frontier, we find that our heuristic provides a reasonable performance bound quickly. After formulating scene-specific SSIM targets and constraining the search, we then leverage generative synthesis to iteratively discover tailored models across a range of complexities specific to each scene's requirements.

\subsection{Network Architecture}
The generated NAS-NeRF architectures, as shown in Table~\ref{tab:architecture_configs}, are much more compact compared to the NeRF~\cite{mildenhall2020nerf} baseline, with up to 23$\times$ fewer parameters for the XXS variants. This compactness leads to substantial speedups of up to 4.7$\times$ frames per second. The architectures also achieve major reductions in computational complexity, with the S, XS, and XXS variants requiring 2.19-4.19$\times$, 4.34-16.8$\times$, and 16.8-22.1$\times$ fewer FLOPs respectively. 

The architectures tend to converge on a minimal coarse field, while exhibiting more diversity in the fine fields, which aligns with the fine field's role in capturing higher frequency details. The network depths of each stage remain fairly consistent for both coarse and fine fields across scenes. However, the channel widths vary substantially, especially in the fine fields. Most coarse fields have uniform stage widths, whereas many fine fields have non-uniform widths with progressive channel expansion across stages. 

The variation in fine field widths and channel growth demonstrates the scene-specific customization enabled by NAS-NeRF's generative architecture search. Notably, the same architecture satisfied constraints for both the Ficus scene at the XXS scale and the Mic scene at the XS scale. This highlights the flexibility of NAS-NeRF in finding tailored architectures across scenes and target performance levels.

\begin{table}[t]
\centering
\begin{subtable}{\textwidth}
\centering
\subcaption*{\textbf{PSNR ↑}}
\begin{tabular}{c|cccccccc}
\toprule
Architecture & Chair & Drums & Ficus & Hotdog & Lego & Materials & Mic & Ship \\
\midrule
\midrule
NeRF & \textbf{31.259} & \textbf{24.607} & \textbf{29.086} & 34.137 & 31.319 & 29.428 & 31.089 & \textbf{27.669} \\
\midrule
NAS-NeRF S & 31.205 & 24.129 & 28.382 & \textbf{34.312} & \textbf{31.720} & \textbf{29.526} & \textbf{31.711} & 26.985 \\
NAS-NeRF XS & 29.423 & 23.215 & 28.100 & 32.307 & 28.266 & 29.135 & 29.313 & 24.876 \\
NAS-NeRF XXS & 27.756 & 20.849 & 24.526 & 28.921 & 25.047 & 27.627 & 27.887 & 22.525 \\
\bottomrule
\end{tabular}
\end{subtable}
\vspace{0.3cm}
\begin{subtable}{\textwidth}
\centering
\subcaption*{\textbf{SSIM ↑}}
\begin{tabular}{c|cccccccc}
\toprule
Architecture & Chair & Drums & Ficus & Hotdog & Lego & Materials & Mic & Ship \\
\midrule
\midrule
NeRF & 0.953 & \textbf{0.912} & \textbf{0.958} & 0.964 & 0.950 & 0.949 & 0.971 & \textbf{0.827} \\
\midrule
NAS-NeRF S & \textbf{0.953} & 0.905 & 0.954 & \textbf{0.966} & \textbf{0.953} & \textbf{0.950} & \textbf{0.975} & 0.814 \\
NAS-NeRF XS & 0.929 & 0.889 & 0.951 & 0.951 & 0.911 & 0.946 & 0.958 & 0.782 \\
NAS-NeRF XXS & 0.904 & 0.847 & 0.908 & 0.921 & 0.848 & 0.932 & 0.947 & 0.757 \\
\bottomrule
\end{tabular}
\end{subtable}
\vspace{0.3cm}
\begin{subtable}{\textwidth}
\centering
\subcaption*{\textbf{LPIPS ↓}}
\begin{tabular}{c|cccccccc}
\toprule
Architecture & Chair & Drums & Ficus & Hotdog & Lego & Materials & Mic & Ship \\
\midrule
\midrule
NeRF & 0.043 & \textbf{0.089} & \textbf{0.039} & 0.039 & 0.027 & 0.042 & 0.033 & \textbf{0.179} \\
\midrule
NAS-NeRF S & \textbf{0.041} & 0.100 & 0.045 & \textbf{0.035} & \textbf{0.026} & \textbf{0.038} & \textbf{0.025} & 0.191 \\
NAS-NeRF XS & 0.071 & 0.124 & 0.046 & 0.062 & 0.057 & 0.044 & 0.052 & 0.266 \\
NAS-NeRF XXS & 0.110 & 0.213 & 0.086 & 0.127 & 0.131 & 0.061 & 0.076 & 0.320 \\
\bottomrule
\end{tabular}
\end{subtable}
\vspace{0.3cm}

\caption{Per-scene quantitative results on the Blender synthetic dataset comparing the generated architectures with the baseline NeRF model.}
\label{tab:blender_results}
\end{table}

\begin{figure}[!ht]
\centering
\includegraphics[width=0.95\linewidth]{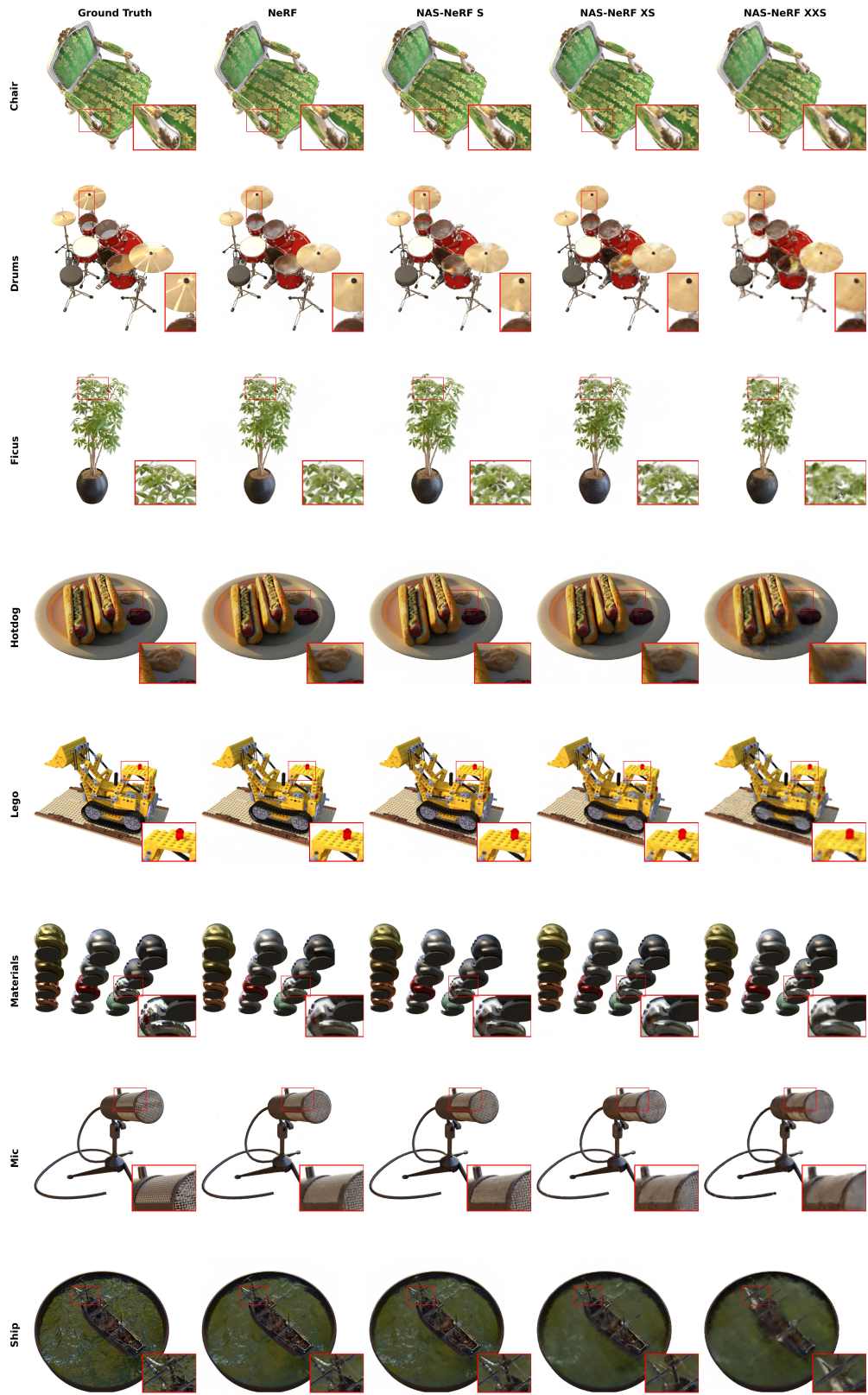}
\caption{Qualitative results of each of our NAS-NeRF architectures evaluated on the corresponding scene it was generated for, along with the ground truth image and NeRF baseline. Red box shows a magnified view of an image patch.}
\label{fig:eval_results_grid}
\end{figure}

\clearpage

\section{Experiments}
We evaluate NAS-NeRF on the Blender synthetic dataset~\cite{mildenhall2020nerf}, comparing SSIM~\cite{wang_image_2004}, PSNR, and LPIPS~\cite{zhang_unreasonable_2018} against the NeRF~\cite{mildenhall2020nerf} baseline. To ensure that all architectures are trained to a comparable level, we scale the number of training iterations proportional to the architecture parameter efficiency ratio $\textit{ER}_{\textit{params}}$ and relative to the number of training iterations required for the baseline NeRF architecture (200k iterations). Note that we define architecture efficiency ratio as $\textit{ER}_{\textit{metric}} = \textit{Metric}(A_{\textit{baseline}}) : \textit{Metric}(A_{\textit{generated}})$.

NAS-NeRF is built on the Nerfstudio framework (v0.2.1)~\cite{tancik_nerfstudio_2023} and all architectures are trained using the default hyperparameters for the \textit{Vanilla NeRF} model (i.e. RAdam optimizer with a learning rate of $5\times 10^{-4}$). To maximize reproducibility and extensibility, we also open source our repo\footnote{\url{https://saeejithnair.github.io/NAS-NeRF/}}.

As seen in Figure~\ref{fig:efficiency} and Table~\ref{tab:blender_results}, our generated architectures demonstrate strong performance given their compact size. The NAS-NeRF S variants match or exceed the baseline NeRF in terms of performance metrics, despite having 2-4$\times$ fewer parameters and running up to 1.93$\times$ faster. The NAS-NeRF XXS variants achieve SSIM scores only 5.3\% lower than baseline NeRF, while requiring 23$\times$ fewer parameters and achieving 4.7$\times$ speedup. Qualitative comparisons in Figure~\ref{fig:eval_results_grid} shows the NAS-NeRF S and XS variants produce rendering quality on par with the baseline, while the smallest XXS models lose some high-frequency details. However, combining our architecture search strategy with newer NeRF techniques for geometric encoding, sampling, and hierarchical scene representation could overcome this limitation and enable high-quality novel view synthesis from ultra-compact models. Overall, the results highlight the ability of NAS-NeRF to find scene-specialized architectures along the complexity-quality spectrum.

\section{Conclusion}
Our work introduces NAS-NeRF, a family of NeRF architectures that are tailored per scene and compute budget. By executing a generative neural architecture search strategy, we show that our method can discover architectures at varying scales of efficiency (NAS-NeRF XXS, XS, and S), thus introducing a way for developers and researchers to systematically make trade offs between performance target quality and architecture complexity, while still remaining near the pareto frontier. By developing NeRF models that achieve competitive performance with baseline NeRF architectures while being a fraction of the size, we hope to open the door to enabling more widespread deployment of novel view synthesis techniques on resource constrained devices like mobile phones and embedded systems. 

{\small
\bibliographystyle{unsrtnat}
\bibliography{Main}
}

\end{document}